\documentclass[journal]{IEEEtran}
\ifCLASSINFOpdf
  \usepackage[pdftex]{graphicx}
    \usepackage{amsmath}
  \usepackage{amssymb}
  \usepackage{bm}
  \usepackage{color}
  \usepackage{array}
  \usepackage{multirow}
  \usepackage{cases}
  \usepackage{algorithm}

\else
\fi
\begin{document}
%
% paper title
% Titles are generally capitalized except for words such as a, an, and, as,
% at, but, by, for, in, nor, of, on, or, the, to and up, which are usually
% not capitalized unless they are the first or last word of the title.
% Linebreaks \\ can be used within to get better formatting as desired.
% Do not put math or special symbols in the title.
\title{Geometry-aware Similarity Learning \\ on SPD Manifolds for Visual Recognition}
%
%
% author names and IEEE memberships
% note positions of commas and nonbreaking spaces ( ~ ) LaTeX will not break
% a structure at a ~ so this keeps an author's name from being broken across
% two lines.
% use \thanks{} to gain access to the first footnote area
% a separate \thanks must be used for each paragraph as LaTeX2e's \thanks
% was not built to handle multiple paragraphs
%

\author{Zhiwu~Huang,~\IEEEmembership{Member,~IEEE,}
        Ruiping~Wang, ~\IEEEmembership{Member,~IEEE,}
        Xianqiu~Li, Wenxian~Liu, \\
        Shiguang~Shan,~\IEEEmembership{Senior Member,~IEEE,}
        Luc~Van Gool, ~\IEEEmembership{Member,~IEEE}
        and ~Xilin~Chen, ~\IEEEmembership{Fellow,~IEEE}% <-this % stops a space
\IEEEcompsocitemizethanks{\IEEEcompsocthanksitem Zhiwu Huang was with the Key Lab of Intelligent Information Processing of Chinese Academy of Sciences (CAS), Institute of Computing Technology (ICT), Beijing, 100190, China. He is now with the Computer Vision Laboratory, Swiss Federal Institute of Technology (ETH), Zurich, 8092, Switzerland. E-mails: zhiwu.huang@vision.ee.ethz.ch.}
\IEEEcompsocitemizethanks{\IEEEcompsocthanksitem Ruiping Wang, Xianqiu Li, Wenxian Liu, Shiguang Shan and Xilin Chen are with the Key Lab of Intelligent Information Processing of Chinese Academy of Sciences (CAS), Institute of Computing Technology (ICT), Beijing, 100190, China. E-mails: \{wangruiping, sgshan, xlchen\}@ict.ac.cn. }
%(Corresponding author: Shiguang Shan.)
\IEEEcompsocitemizethanks{\IEEEcompsocthanksitem Luc Van Gool is with the Computer Vision Laboratory, Swiss Federal Institute of Technology (ETH), Zurich, 8092, Switzerland. E-mails: vangool@vision.ee.ethz.ch.}}

\maketitle

% As a general rule, do not put math, special symbols or citations
% in the abstract or keywords.
\begin{abstract}
Symmetric Positive Definite (SPD) matrices have been widely used for data representation in many visual recognition tasks. The success mainly attributes to learning discriminative SPD matrices with encoding the Riemannian geometry of the underlying SPD manifold. In this paper, we propose a geometry-aware SPD similarity learning (SPDSL) framework to learn discriminative SPD features by directly pursuing manifold-manifold transformation matrix of column full-rank. Specifically, by exploiting the Riemannian geometry of the manifold of fixed-rank Positive Semidefinite (PSD) matrices, we present a new solution to reduce optimizing over the space of column full-rank transformation matrices to optimizing on the PSD manifold which has a well-established Riemannian structure. Under this solution, we exploit a new supervised SPD similarity learning technique to learn the transformation by regressing the similarities of selected SPD data pairs to their ground-truth similarities on the target SPD manifold. To optimize the proposed objective function, we further derive an algorithm on the PSD manifold. Evaluations on three visual classification tasks show the advantages of the proposed approach over the existing SPD-based discriminant learning methods.
\end{abstract}

% Note that keywords are not normally used for peerreview papers.

% For peer review papers, you can put extra information on the cover
% page as needed:
% \ifCLASSOPTIONpeerreview
% \begin{center} \bfseries EDICS Category: 3-BBND \end{center}
% \fi
%
% For peerreview papers, this IEEEtran command inserts a page break and
% creates the second title. It will be ignored for other modes.
\IEEEpeerreviewmaketitle

\section{Introduction}
% The very first letter is a 2 line initial drop letter followed
% by the rest of the first word in caps.
%
% form to use if the first word consists of a single letter:
% \IEEEPARstart{A}{demo} file is ....
%
% form to use if you need the single drop letter followed by
% normal text (unknown if ever used by the IEEE):
% \IEEEPARstart{A}{}demo file is ....
%
% Some journals put the first two words in caps:
% \IEEEPARstart{T}{his demo} file is ....
%
% Here we have the typical use of a "T" for an initial drop letter
% and "HIS" in caps to complete the first word.
Recently, Symmetric Positive Definite (SPD) matrices of real numbers appear in many branches of computer vision. Examples include region covariance matrices for pedestrian detection \cite{tuzel2006region,tuzel2008pedestrian} and texture categorization \cite{harandi2012sparse,jay2013kernel,harandi2014manifold}, joint covariance descriptor for action recognition \cite{Hussein2013,harandi2014manifold}, diffusion tensors for DT image segmentation \cite{pennect2006aid,arsigny2007led,jay2013kernel} and image set based covariance matrix for video face recognition \cite{wang2012covariance,vemulapalli2013kernel,lu2013image}. Due to the effectiveness of measuring data variations, such SPD features have been shown to provide powerful representations for images and videos.
% in computer vision.

However, such advantages of the SPD matrices often accompany with the challenge of their non-Euclidean data structure which underlies a specific Riemannian manifold \cite{pennect2006aid,arsigny2007led}. Applying the Euclidean geometry directly to SPD matrices often results in poor performances and undesirable effects, such as the swelling of diffusion tensors in the case of SPD matrices \cite{arsigny2006log,penec2006rieman}. To overcome the drawbacks of the Euclidean representation, recent works \cite{penec2006rieman,arsigny2007led,sra2011sdiv} have introduced Riemannian metrics, e.g., Affine-Invariant metric \cite{pennect2006aid}, Log-Euclidean metric \cite{arsigny2007led}, to encode the Riemannian geometry of SPD manifold properly.
%classical Riemannian metrics

\begin{figure}[t]
\begin{center}
   \includegraphics[width=0.9\linewidth]{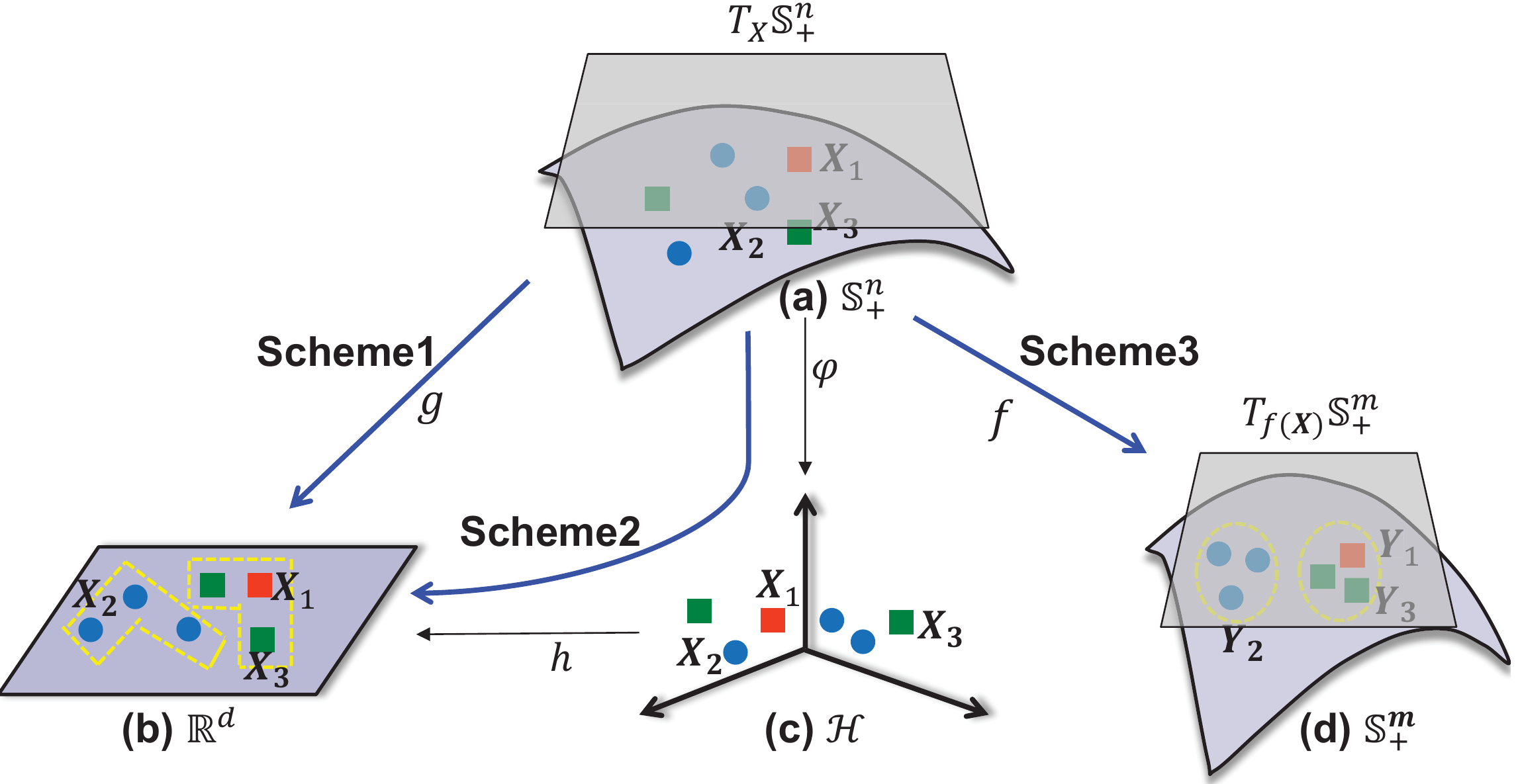}
\end{center}
   \caption{Three different learning schemes for SPD features. The first one (a)$\rightarrow$(b) is to firstly flatten the original manifold $Sym_{+}^{n}$ by tangent space approximation and then learn a map $g$ to a discriminative Euclidean space $\mathbb{R}^d$. The second one (a)$\rightarrow$(c)$\rightarrow$(b) is to firstly embed $Sym_{+}^{n}$ with an implicit map $\varphi$ into an RKHS $\mathcal{H}$ and then learn a mapping $h$ to a more discriminative Euclidean space $\mathbb{R}^d$. The last one (a)$\rightarrow$(d) aims to learn a map $f$ from the original SPD manifold $Sym_{+}^{n}$ to a more discriminative SPD manifold $Sym_{+}^{m}$. Here, $\bm{X} \in Sym_{+}^{n}$ and $f(\bm{X}) \in Sym_{+}^{m}$ are the SPD matrices, $T_{X}Sym_{+}^{n}$ and $T_{f(X)}Sym_{+}^{m}$ are the tangent spaces.}
\label{fig:long}
\label{Fig1}
\end{figure}

By applying these classical Riemannian metrics, a couple of works attempt to extend Euclidean algorithms to work on manifolds of SPD matrices for learning more discriminative SPD matrices or their vector-forms. To this end, several studies exploit effective methods on one SPD manifold by either flattening it via tangent space approximation \cite{tuzel2008pedestrian,tosato2010region,carreira2012sem,sanin2013spatio} (See Fig.\ref{Fig1} (a)$\rightarrow$(b)) or mapping it into a high dimensional Reproducing Kernel Hilbert Space (RKHS) \cite{harandi2012sparse,wang2012covariance,jay2013kernel, harandi2014bregman, faraki2015approximate, harandi2016sparse, minh2014nips} (See Fig.\ref{Fig1} (a)$\rightarrow$(c)$\rightarrow$(b)). Obviously, both of the two families of methods inevitably distort the geometrical structure of the original SPD manifold due to the procedure of mapping the manifold into a flat Euclidean space or a high dimensional RKHS. Therefore, the two learning schemes would lead to sub-optimal solutions for the problem of discriminative SPD matrix learning.

\vspace{1em}

To more faithfully respect the original Riemannian geometry, another kind of SPD-based discriminant learning methods\footnote{Several related ideas were introduced in \cite{liu2003optimal,jung2012analysis,huang2015projection} for the dimensionality reduction or optimization on different types of Riemannian manifolds.} \cite{harandi2014manifold,huang2015leml} aims to pursue a column full-rank transformation matrix mapping the original SPD manifold to a more discriminative SPD manifold, as shown in Fig.\ref{Fig1} (a)$\rightarrow$(d). However, as directly learning the manifold-manifold transformation matrix is hard, the work \cite{harandi2014manifold} alternatively decomposes it to the product of an orthonormal matrix with a matrix in GL$(n)$, and requires the employed Riemannian metrics to be affine invariant. By doing so, optimizing the manifold-manifold transformations is equivalent to optimizing over orthonormal projections. Although the additional requirement simplifies the optimization of the transformation, this has not only reduced the original solution space but also inevitably excluded all non-affine invariant Riemannian metrics such as the well-known Log-Euclidean metric, which has proved to be much more efficient than Affine-Invariant metric \cite{arsigny2007led}. While the work \cite{huang2015leml} exploited the Log-Euclidean metric under the same scheme, it actually attempts to learn a tangent map, which implicitly approximate the tangent space and hence introduces some distortions of the true geometry of SPD manifolds.

In this paper, also under the last scheme (see Fig.\ref{Fig1} (a)$\rightarrow$(d)), we propose a new geometry-aware SPD similarity learning (SPDSL) framework to open a broader problem domain of learning discriminative SPD features by exploiting either affine invariant or non-affine invariant Riemannian metrics on SPD manifolds. To realize the SPDSL framework, there are three main contributions in this work:
\begin{itemize}
  \item By exploiting the Riemannian geometry of the manifold of fixed-rank Positive Semidefinite (PSD) matrices, our SPDSL framework provides a new solution to directly learn the manifold-manifold transformation matrix. As no additional constraint is required, the optimal transformation will be pursued in a favorable solution space, enabling a wide range of well-established Riemannian metrics to work as well.
  \item To fulfill the solution, a new supervised SPD similarity learning technique is proposed to learn the transformation by regressing the similarities of selected SPD pairs to the target similarities on the resulting SPD manifold.
  \item We derive an optimization approach which exploits the classical Riemannian Conjugate Gradient (RCG) algorithm on the PSD manifold to optimize the proposed objective function.
\end{itemize}

\section{Background}
\label{sec2}

Let $Sym_{n}=\{\bm{H}: \bm{H}^T=\bm{H}\}$ be a set of real, symmetric matrices of size $n \times n$ and $Sym^{+}_{n}=\{\bm{X} \in Sym_{n}: \bm{\omega}^T\bm{X}\bm{\omega} \succ 0, \forall \bm{\omega} \in \mathbb{R}^{n}, \bm{\omega} \neq 0\}$ be a set of SPD matrices. The mapping space $Sym_{n}$ is endowed with usual Euclidean metric (i.e., inner product) $ \langle \bm{H}_1, \bm{H}_2  \rangle = Tr(\bm{H}_2^T \bm{H}_1)$. As noted in \cite{pennect2006aid,arsigny2007led}, the set of SPD matrices $Sym^{+}_{n}$ is an open convex subset of $Sym_{n}$. Thus, the tangent space to $Sym^{+}_{n}$ at any SPD matrix in it can be identified with the set $Sym_{n}$. A smoothly-varying family of inner products on each tangent space is known as Riemannian metric, endowing which the space of SPD matrices $Sym^{+}_{n}$ would yield a Riemannian manifold. With such Riemannian metric, the geodesic distance between two elements $\bm{X}_1,\bm{X}_2$ on the SPD manifold is generally measured by $\langle \log_{\bm{X}_1}(\bm{X}_2), \log_{\bm{X}_1}(\bm{X}_2)  \rangle_{\bm{X}_1}$. Several Riemannian metrics and divergences have been proposed to equip SPD manifolds. For example, Affine-Invariant metric \cite{pennect2006aid}, Stein divergence \cite{sra2012new}, Jeffereys divergence \cite{harandi2014bregman} are designed to be invariant to affine transformation. That is, for any $\bm{M} \in GL(n)$ (i.e., the group of real invertible $n\times n$ matrices), the metric function $\delta_A$ has the property $\delta_A^2(\bm{X}_1,\bm{X}_2)=\delta_A^2(\bm{M}\bm{X}_1\bm{M}^T,\bm{M}\bm{X}_2\bm{M}^T)$. In contrast, Log-Euclidean metric\cite{arsigny2007led}, Cholesky distance \cite{dryden2009non} and Power-Euclidean metric \cite{dryden2009non} are not affine invariant. Among these metrics, only Affine-Invariant metric \cite{pennect2006aid} and Log-Euclidean metric \cite{arsigny2007led} define a true geodesic distance on the SPD manifold \cite{jay2013kernel}. In addition, the Stein divergence are also widely used due to its favorable properties and high performances in visual recognition tasks \cite{sra2012new}. Therefore, this paper focuses on studying such three representative Riemannian metrics.

\vspace{1em}

\noindent \textbf{Definition 1.} \emph{By defining the inner product in the tangent space at the SPD point $\bm{X}_1$ on the SPD manifold as $\langle \bm{H}_1, \bm{H}_2  \rangle_{\bm{X}_1} = \langle \bm{X}_1^{-1/2}\bm{H}_1\bm{X}_1^{-1/2}, \bm{X}_1^{-1/2} \bm{H}_2 \bm{X}_1^{-1/2} \rangle$ and the logarithmic maps as $\log_{\bm{X}_1}(\bm{X}_2)=\bm{X}_1^{1/2}\log(\bm{X}_1^{-1/2}\bm{X}_2\bm{X}_1^{-1/2})\bm{X}_1^{1/2}$, the geodesic distance between two SPD matrices $\bm{X}_1,\bm{X}_2$ on the SPD manifold is induced by Affine-Invariant metric (AIM) as}
\begin{equation}
\begin{aligned}
\delta_a^2(\bm{X}_1,\bm{X}_2)  = \|\log(\bm{X}_1^{-1/2}\bm{X}_2\bm{X}_1^{-1/2})\|_{\mathcal{F}}^2.
\label{Eq1}
\end{aligned}
\end{equation}

\noindent \textbf{Definition 2.} \emph{The approximated geodesic distance between two SPD matrices $\bm{X}_1,\bm{X}_2$ on the SPD manifold is defined by using Stein divergence as}
\begin{equation}
\begin{aligned}
\delta_a^2(\bm{X}_1,\bm{X}_2) =  \ln \det \left( \frac{\bm{X}_1+\bm{X}_2}{2}\right)-\frac{1}{2}\ln\det(\bm{X}_1\bm{X}_2).
\label{Eq1.0}
\end{aligned}
\end{equation}

\noindent \textbf{Definition 3.} \emph{By defining the inner product in the \mbox{tangent} space at the SPD point $\bm{X}_1$ on the SPD manifold as $\langle \bm{H}_1, \bm{H}_2  \rangle_{\bm{X}_1} = \langle \text{D}\log(\bm{X}_1)[\bm{H}_1], \text{D}\log(\bm{X}_1)[\bm{H}_2] \rangle$ ($\text{D}\log(\bm{X}_1)[\bm{H}]$ denotes the directional derivative) and the logarithmic maps as $\log_{\bm{X}_1}(\bm{X}_2)= \text{D}^{-1} \log(\bm{X}_1)[\log(\bm{X}_2)-\log(\bm{X}_1)]$, the geodesic distance between two SPD matrices $\bm{X}_1,\bm{X}_2$ is derived by Log-Euclidean metric (LEM) as}
\begin{equation}
\begin{aligned}
\delta_l^2(\bm{X}_1,\bm{X}_2) = \|\log(\bm{X}_1)-\log(\bm{X}_2)\|_{\mathcal{F}}^2.
\label{Eq2}
\end{aligned}
\end{equation}

\section{Proposed Approach}
\label{sec3}

In this section, we first propose a new solution of Riemannian geometry-aware dimensionality reduction for SPD matrices, and then present our supervised SPD similarity learning method under the solution. Finally, we give a detailed description of our developed optimization algorithm.
% to solve this problem.

%------------------------------------------------------------------------
\subsection{Riemannian Geometry-aware Dimensionality Reduction on SPD manifolds}

Given a set of SPD matrices $\bm{X}=\{\bm{X}_1,\ldots,\bm{X}_N\}$, where each matrix $\bm{X}_i \in Sym^{+}_{n}$, and a transformation $\bm{W}  \in \mathbb{R}^{n \times m}$ ($m < n$) is pursued for mapping the original SPD manifold $Sym^{+}_{n}$ to a lower-dimensional SPD manifold $Sym^{+}_{m}$. Formally, this procedure attempts to learn the parameter $\bm{W}$, of a mapping in the form $f: Sym^{+}_{n} \times \mathbb{R}^{n \times m} \rightarrow Sym^{+}_{m}$, which is defined as:
\begin{equation}
f(\bm{X}_i,\bm{W}) = \bm{W}^T\bm{X}_i\bm{W}.
\label{Eq3}
\end{equation}
To ensure the resulting mapping yields a valid SPD manifold $Sym^{+}_{m} \ni \bm{W}^T\bm{X}_i\bm{W}$ $\succ 0$, the manifold-manifold transformation $W$ is basically required to be a \mbox{column} full-rank matrix $\bm{W} \in \mathbb{R}_{*}^{n \times m}$.

Since the solution space is a non-compact Stiefel manifold $\mathbb{R}^{n \times m}_{*}$ where the distance function has no upper bound, directly optimizing on the manifold is infeasible. Fortunately, the conjugates (taking the form of $\bm{W}\bm{W}^T$) of column full-rank matrices span a compact manifold $Sym^{+}_{n}(m)$ of Positive Semidefinite (PSD) matrices, which is a quotient space of $\mathbb{R}^{n \times m}_{*}$ and owns a well-established Riemannian structure. In contrast, by additionally assuming the transformation $\bm{W}$ to be orthogonal as done in \cite{harandi2014manifold}, Eqn.\ref{Eq3} could be optimized on compact Stiefel manifold, which is a subset of the non-compact Stiefel manifold $\mathbb{R}^{n \times m}_{*}$. Further, for the affine invariant metrics (e.g., AIM), optimizing on Stiefel manifold can be reduced to optimizing over Grassmannian \cite{harandi2014manifold}. However, such orthogonal solution space is smaller than the original solution space $\mathbb{R}^{n \times m}_{*}$, making the optimization theoretically yield suboptimal solution of $\bm{W}$. Thus, we choose to optimization on the PSD manifold to search the optimal solution of $\bm{W}$ in a more faithful way. Now, we need to study the geometry of the PSD manifold $Sym^{+}_{n}(m)$.

For all orthogonal matrices $\bm{O}$ of size $m \times m$, the map $\bm{W} \rightarrow \bm{WO}$ leaves $\bm{W}\bm{W}^T$ unchanged. This property of $\bm{W}$ results in the equivalence class of the form $[\bm{W}] =\{\bm{WO}| \bm{O} \in \mathbb{R}^{m \times m}, \bm{O}^T\bm{O}=\bm{I}_m\}$, and \mbox{yields} a one-to-one correspondence with the rank-$m$ PSD matrix $\bm{Q}=\bm{W}\bm{W}^T \in Sym^{+}_{n}(m)$. By quotienting this equivalence relation out, the set of rank-$m$ PSD matrices $Sym^{+}_{n}(m)$ is reduced to the quotient of the manifold $\mathbb{R}_{*}^{n \times m}$ by the orthogonal group $\mathcal{O}(m)=\{\bm{O} \in \mathbb{R}^{m \times m} | \bm{O}^T\bm{O}=\bm{I}_m\}$, i.e., $Sym^{+}_{n}(m)=\mathbb{R}_{*}^{n \times m}/\mathcal{O}(m)$.
With the studied relationship between $Sym^{+}_{n}(m)$ and $\mathbb{R}_{*}^{n \times m}$, the function $\phi: Sym^{+}_{n}(m) \rightarrow \mathbb{R}: \bm{\bm{Q}} \mapsto \phi(\bm{\bm{Q}})$ is able to derive the function $g: \mathbb{R}_{*}^{n \times m} \rightarrow \mathbb{R}: \bm{W} \mapsto g(\bm{W})$ defined as $g(\bm{W})=\phi(\bm{W}\bm{W}^T)$. Here, $g$ is defined in the total space $\mathbb{R}_{*}^{n \times m}$ and descends as a well-defined function in the quotient manifold $Sym^{+}_{n}(m)$. Therefore, optimizing over the total space $\mathbb{R}_{*}^{n \times m}$ is reduced to optimizing on the PSD manifold $Sym^{+}_{n}(m)$, which is well-studied in several works \cite{absil2008optimization,bonnabel2009riemannian,journee2010lowrank,meyer2011regression}.
Note that, as each element $\bm{Q}=\bm{W}\bm{W}^T$ on the PSD manifold is simply parameterized by $\bm{W}$, optimizing on the manifold actually deals directly with $\bm{W}$.
To more easily understand this point, one can take the well-known Grassmann manifold as an analogy, where each element can be similarly represented by the equivalence class $[\bm{W}]$ or the projection matrix $\bm{W}\bm{W}^T$ (here, $\bm{W}^T\bm{W}=\bm{I}$), and the optimization on it directly seeks the solution of $\bm{W}$.

It can be further proven that the quotient $Sym^{+}_{n}(m)$ presents the structure of a Riemannian manifold \cite{absil2008optimization}. As a result, endowing the total space $\mathbb{R}_{*}^{n \times m}$ with the usual Riemannian structure of a Euclidean space (i.e., the inner product $ \langle \bm{H}_1, \bm{H}_2  \rangle = Tr(\bm{H}_2^T \bm{H}_1)$), a Riemannian structure for the quotient space $Sym^{+}_{n}(m)$ follows. The inner product occurs on the tangent space $T_W$ of the manifold $\mathbb{R}_{*}^{n \times m}$. In the case of the manifold $Sym^{+}_{n}(m)$, the corresponding tangent space is decomposed into two orthogonal subspaces, the vertical space $\mathcal{V}_W=\{\bm{W}\bm{\Omega} | \bm{\Omega} \in \mathbb{R}^{n \times m}, \bm{\Omega}^T=-\bm{\Omega}\}$ and the horizontal space $\mathcal{H}_W = \{\bm{H} \in T_W | \bm{H}^T\bm{W}=\bm{W}^T\bm{H}\}$, to achieve the inner product $\langle \bm{H}_1, \bm{H}_2  \rangle$. This Riemannian metric facilitates several classical optimization techniques such as Riemannian Conjugate Gradient (RCG) algorithm \cite{absil2008optimization} working on the PSD manifold $Sym^{+}_{n}(m)$. As for more detailed background on the Riemannian geometry of the PSD manifold, please refer to the works \cite{absil2008optimization,journee2010lowrank}.

By exploiting the Riemannian geometry of the fixed-rank PSD manifold $Sym^{+}_{n}(m)$, we here open up the possibility of directly pursuing an optimal column full-rank manifold-manifold transformation matrix to solve the problem of dimensionality reduction on SPD features.

%------------------------------------------------------------------------
\subsection{Supervised SPD similarity learning}

As studied before, under the proposed framework of dimensionality reduction on SPD features, a target SPD manifold $Sym^{+}_{m}$ of lower dimensionality can be derived. On the new SPD manifold $Sym^{+}_{m}$, the geodesic distance between the two original SPD points $\bm{X}_i, \bm{X}_j$ is achieved by:
\begin{equation}
 \hat{\delta}^2(\bm{X}_i, \bm{X}_j)=\delta^2(f(\bm{X}_i, \bm{W}), f(\bm{X}_j, \bm{W})),
 \label{Eq4}
 \end{equation}
where $f(\bm{X}_i, \bm{W})$ is the manifold-manifold transformation computed by Eqn.\ref{Eq3}, $\delta$ can be the geodesic distance induced by the commonly-used affine or non-affine invariant Riemannian metrics Eqn.\ref{Eq1}, Eqn.\ref{Eq1.0} and Eqn.\ref{Eq2}.
% showing in Table \ref{tab1}.

In this paper, we are focusing on the problem of supervised SPD similarity learning for more robust visual classification tasks where SPD features have shown great power. Formally, for each SPD matrix $\bm{X}_i \in Sym^{+}_{n}$, we define its class indicator vector: $\bm{y}_i=[0, \ldots, 1, \ldots, 0] \in \mathbb{R}^c$, where the $k$-th entry being 1 and other entries being 0 indicates that $\bm{X}_i$ belongs to the $k$-th class of $c$ classes in total. As discriminant learning techniques developed in Euclidean space, we assume that prior knowledge is known regarding the distances between pairs of SPD points on the new SPD manifold $Sym^{+}_{m}$. Let's take the similarity or dissimilarity between pairs of SPD points into account: two SPD points are similar if the similarity based on the geodesic distance between them on the new manifold is larger, while two SPD points are dissimilar if their similarity is smaller.

Given a set of the similarity constraints, our goal is to learn the manifold-manifold transformation matrix $\bm{W}$ that parameterizes the similarities of SPD points on the target SPD manifold $Sym^{+}_{m}$. To this end, we exploit the supervised criterion of centered kernel target alignment \cite{cristianini2001on,cortes2012alogrithm,Yger2015supevised} to learn discriminative features on the SPD manifold by regressing the similarities of selected sample pairs to the target similarities. Formally, our supervised SPD similarity learning (SPDSL) approach is to maximize the following objective function:
\begin{equation}
\begin{aligned}
\mathcal{J}(\bm{W}) = \frac{\langle \bm{U}\bm{G}\circ k({\bm{W}})\bm{U}, \bm{G}\circ(\bm{Y}\bm{Y}^{T}) \rangle_{\mathcal{F}}}{\|\bm{U}\bm{G}\circ k({\bm{W}})\bm{U}\|_{\mathcal{F}}}, s.t. \bm{W} \in \mathbb{R}_{*}^{n \times m},
\label{Eq9}
\end{aligned}
\end{equation}
where $\langle \cdot \rangle_{\mathcal{F}}$ and $\|\cdot\|_{\mathcal{F}}$ are Frobenius inner product and norm respectively.
The elements of matrix $k(\bm{W})$ encodes the similarities of SPD data while the elements of $\bm{Y}\bm{Y}^{T}$ presents the ground-truth similarities of the involved SPD points. The matrix $\bm{G}$ is used to select the pairs of SPD points when the corresponding elements are 1.
The matrix $\bm{U}=\bm{I}_N-\frac{\bm{1}_N \bm{1}_N^T}{N}$ is employed for centering the data similarity matrix $k(\bm{W})$ and the similarity matrix $\bm{Y}\bm{Y}^{T}$ on labels. $N$ is the number of samples, $\bm{I}_N$ is the identity matrix of size $N \times N$, $\bm{1}_N$ is the vector of size $N$ with all entries being ones, $\bm{Y}=[\bm{y}_1, \ldots, \bm{y}_N]^T$ is here supposed to be centered, i.e., $\bm{U} (\bm{Y}\bm{Y}^{T})\bm{U}\rightarrow\bm{Y}\bm{Y}^{T}$, for simplicity. In the following, we will give the formulations of the two matrices $k({\bm{W}})$ and $\bm{G}$ in more details.

More specifically, the employed matrix $k({\bm{W}})$ in Eqn.\ref{Eq9} encodes the similarity between each pair of SPD points $(\bm{X}_i, \bm{X}_j)$ on the SPD manifold $Sym^{+}_{n}$, which takes a form as:
\begin{equation}
 k_{ij}({\bm{W}})=\exp(-\beta \hat{\delta}^2(\bm{X}_i, \bm{X}_j)),
 \label{Eq6}
\end{equation}
where $\hat{\delta}^2(\bm{X}_i, \bm{X}_j)$ is computed by Eqn.\ref{Eq4}, $\beta$ is typically set as $\beta=\frac{1}{\sigma^2}$, $\sigma$ is empirically set to mean of distances of the original training sample pairs. Actually, the function Eqn.\ref{Eq6} takes a form of Gaussian kernel function. However, as the objective function Eqn.\ref{Eq9} can be expressed as sum of the similarity regression results of selected sample pairs, the function Eqn.\ref{Eq6} just serves as a tool to encode the similarities and is thus not necessarily positive definite (PD).

In practical application, the computational burden of handling the full kernel matrix $k({\bm{W}})$ on the SPD manifold scales quadratically with the size of training SPD data. To address this problem, we exploit the idea of Graph Embedding technique \cite{yan2007graph} to select a limited number of data pairs to construct a {sparse kernel matrix} (non PD) with a large number of elements being zero. With this idea in mind, the matrix $\bm{G}$ is defined to select the pairs of SPD points for SPD similarity learning. By employing it, $\bm{G}\circ k({\bm{W}})$ can be regarded as the {sparse kernel matrix}, where the operation $\circ$ denotes Hadamard product and the matrix $\bm{G}=\bm{G}_w+\bm{G}_b$. Here, $\bm{G}_w$ and $\bm{G}_b$ are defined as:
\begin{align}
\bm{G}_w(i,j)& =
\begin{cases}
1, & \text{if $\bm{X}_i \in N_w(\bm{X}_j)$ or $\bm{X}_j \in N_w(\bm{X}_i)$}\\
0, & \text{otherwise},
\end{cases}\label{Eq7}\\
\bm{G}_b(i,j)& =
\begin{cases}
1, & \text{if $\bm{X}_i \in N_b(\bm{X}_j)$ or $\bm{X}_j \in N_b(\bm{X}_i)$}\\
0, & \text{otherwise},
\end{cases}
\label{Eq8}
\end{align}
where $N_w(\bm{X}_i)$ is the set of $v_w$ nearest neighbors of $\bm{X}_i$ that share the same class label as $y_i$, and $N_b(\bm{X}_i)$ is the set of $v_b$ nearest neighbors of $\bm{X}_i$ with different class labels from $y_i$. According to the theory of Graph Embedding \cite{yan2007graph}, the within-class similarity graph $\bm{G}_w$ and the between-class dissimilarity graph $\bm{G}_b$ respectively defined in Eqn.\ref{Eq7} and Eqn.\ref{Eq8} can encode the local geometrical structure of the space of the processing data. Thus, in addition to speeding up the discriminant learning on the SPD features, exploiting the Graph Embedding technique can not only learn the discriminative information of SPD data but also characterize the local Riemannian geometry of the underlying SPD manifold. The efficiency and effectiveness of the proposed discriminant learning approach working on SPD manifolds will be further studied in the experimental part.

%-------------------------------------------------------------------------
\subsection{Riemannian Conjugate Gradient Optimization}

As discussed before, optimizing in the solution space $\mathbb{R}_{*}^{n \times m}$ of the column full-rank transformation matrices in our objective function can be reduced to optimizing on the Riemannian manifold of rank-$m$ PSD matrices, $Sym^{+}_{n}(m)$. Therefore, in this section, we exploit the Riemannian Conjugate Gradient (RCG) algorithm \cite{absil2008optimization} to optimize our objective function $\mathcal{J}(\bm{W})$ in Eqn.\ref{Eq9} by deriving its corresponding gradient on the PSD manifold $Sym^{+}_{n}(m)$.

\begin{algorithm}[t]
\caption{Optimization algorithm}
\textbf{Input}: The initial matrix $\bm{W}_{0}$\\
1. $\bm{H}_{0} \leftarrow 0, \bm{W} \leftarrow \bm{W}_{0}$.\\
2. \textbf{Repeat}\\
3. \quad $\bm{H}_k \leftarrow -\nabla_{\bm{W}} J(\bm{W}_k)+\eta \tau(\bm{H}_{k-1},\bm{W}_{k-1},\bm{W}_k)$.\\
4. \quad Line search along the geodesic $\gamma$ with the direction $\bm{H}_k$ from $\bm{W}_{k-1}=\gamma(k-1)$ to find $\bm{W}_{k}=\arg\min_{\bm{W}} \mathcal{J}(\bm{W})$.\\
5. \quad $\bm{H}_{k-1} \leftarrow \bm{H}_k$, $\bm{W}_{k-1} \leftarrow \bm{W}_k$.\\
8. \textbf{Until} convergence\\
\textbf{Output}: The optimized matrix $\bm{W}$
\label{Alg1}
\end{algorithm}

As the Conjugate Gradient algorithm developed in Euclidean space, the RCG algorithm on Riemannian manifolds is an iterative procedure.  As given in Algorithm\ref{Alg1}, an outline for the iterative part of the algorithm goes as follows: at the $k$-th iteration, find $\bm{W}_k$ by searching the minimum of $\mathcal{J}$ along the geodesic in the direction $\bm{H}_{k-1}$ from $\bm{W}_{k-1}$, compute the Riemannian \mbox{gradient} $\nabla_W \mathcal{J}(\bm{W}_k)$ at this point, choose the new search direction by $\bm{H}_k = -\nabla_W \mathcal{J}(\bm{W}_k)+\eta \tau(\bm{H}_{k-1},\bm{W}_{k-1},\bm{W}_k)$ and iterate until convergence. In the procedure, the Riemannian
\mbox{gradient} $\nabla_W \mathcal{J}(\bm{W}_k)$ can be easily approximately from its corresponding Euclidean gradient $D_W \mathcal{J}(\bm{W}_k)$ by the computation $\nabla_W \mathcal{J}(\bm{W}_k)=D_W \mathcal{J}(\bm{W}_k)-\bm{W}_k\bm{W}_k^T D_W \mathcal{J}(\bm{W}_k)$, and the operation $\tau(\bm{H}_{k-1}, \bm{W}_{k-1}, \bm{W}_{k})$ is the parallel transport of tangent vector $\bm{H}_{k-1}$ from $\bm{W}_{k-1}$ to $\bm{W}_{k}$. For more details, we refer readers to \cite{absil2008optimization,harandi2014manifold}.

As for now, we just need to compute the Euclidean gradient for our objective function $\mathcal{J}(\bm{W})$ in Eqn.\ref{Eq9}. As the Euclidean gradient $D_W\mathcal{J}(\bm{W})$ and its corresponding directional derivatives are related with the following equality:
\begin{equation}
D_W\mathcal{J}(\bm{W}) [\dot{\bm{W}}]=\langle D_W\mathcal{J}(\bm{W}), \dot{\bm{W}} \rangle.
\label{Eq10}
\end{equation}
By employing the basic rule and standard properties of the directional derivatives, $D_W\mathcal{J}(\bm{W}) [\dot{\bm{W}}]$ can be derived by:
\begin{equation}
\begin{aligned}
&D_W\mathcal{J}(\bm{W}) [\dot{\bm{W}}] \\
&= \frac{\langle \bm{U}\bm{G}\circ D_W k(\bm{W})[\dot{\bm{W}}]\bm{U}, \bm{G}\circ(\bm{Y}\bm{Y}^{T}) \rangle_{\mathcal{F}} \|\mathcal{L}\|_{\mathcal{F}}}{\|\mathcal{L}\|_{\mathcal{F}}^2}\\
-& \frac{\langle \mathcal{L}, \bm{G}\circ(\bm{Y}\bm{Y}^{T}) \rangle_{\mathcal{F}} \langle \frac{\mathcal{L}}{\|\mathcal{L}\|_{\mathcal{F}}}, \bm{U}\bm{G}\circ D_W k(\bm{W})[\dot{\bm{W}}]\bm{U} \rangle_{\mathcal{F}}}{\|\mathcal{L}\|_{\mathcal{F}}^2} \\
&= \langle D_W k(\bm{W})[\dot{\bm{W}}], \bm{U} \left ( \frac{\bm{G}\circ(\bm{Y}\bm{Y}^{T})}{\|\mathcal{L}\|_{\mathcal{F}}}-
\frac{\mathcal{J}(\bm{W})\mathcal{L}}{\|\mathcal{L}\|_{\mathcal{F}}^2} \right) \bm{U} \rangle_{\mathcal{F}},
\label{Eq12}
\end{aligned}
\end{equation}
where $\mathcal{L}=\bm{U}\bm{G}\circ k({\bm{W}})\bm{U}$, $\langle \cdot \rangle_{\mathcal{F}}$ indicates Frobenius inner product, $\|\cdot\|_{\mathcal{F}}$ denotes Frobenius norm.

Accordingly, the key issue in Eqn.\ref{Eq12} is to estimate $D_W k(\bm{W})$, where $ k(\bm{W})$ is formulated by Eqn.\ref{Eq6}. When $\delta$ in Eqn.\ref{Eq4} is the geodesic distance of AIM defined in Eqn.\ref{Eq1}, the Euclidean gradient of $k(\bm{W})$ can be derived as:
\begin{equation}
\begin{aligned}
& D_W k_{ij}(\bm{W}) \\ & =  -4 \beta k_{ij}(\bm{W})(\bm{B}_i\bm{\hat{X}}_i^{-1}-\bm{B}_j\bm{\hat{X}}_j^{-1}) \log(\bm{\hat{X}}_j^{-\frac{1}{2}}\bm{\hat{X}}_i\bm{\hat{X}}_j^{-\frac{1}{2}}),
\label{Eq11}
\end{aligned}
\end{equation}
where $\bm{B}_i= \bm{X}_i\bm{W}$, $\bm{\hat{X}}_i = \bm{W}^T\bm{X}_i\bm{W} \in Sym^{+}_m$.

For other affine invariant metrics such as Stein divergence \cite{sra2012new}, the corresponding Euclidean gradient of $k(\bm{W})$  with the geodesic distance function $\delta$ being defined in Eqn.\ref{Eq1.0} can be computed by:
\begin{equation}
\begin{aligned}
& D_W k_{ij}(\bm{W}) \\ & =  -\beta k_{ij}(\bm{W})((\bm{B}_i+\bm{B}_j)\bm{A}_{ij}^{-1}-\bm{B}_i\bm{\hat{X}}_i^{-1}-\bm{B}_j\bm{\hat{X}}_j^{-1}),
\label{Eq19}
\end{aligned}
\end{equation}
where $\bm{A}_{ij}=\bm{W}^T\frac{\bm{X}_i+\bm{X}_j}{2}\bm{W}$, and hence be able to work in our new proposed framework.

When endowing the SPD manifold with the non-affine invariant metric LEM, it \mbox{seems} not easy to calculate the Euclidean gradient of $D_W k(\bm{W})$ due to the matrix logarithms in it. Thus, we need to study the problem of the computation of the Euclidean gradient for the LEM case in the following.

First, we decompose the derivative of LEM w.r.t. $\bm{W}$ into three derivatives with the trace form $Tr(\cdot)$:
\begin{equation}
\begin{aligned}
&D_W (\|\log(\bm{W}^T\bm{X}_i\bm{W})-\log(\bm{W}^T\bm{X}_j\bm{W})\|_F^2)= \\
& D_W(Tr(\log^2(\bm{W}^T\bm{X}_i\bm{W})) +D_W(Tr(\log^2(\bm{W}^T\bm{X}_j\bm{W}))\\
& -2D_W(Tr(\log(\bm{W}^T\bm{X}_i\bm{W})\log(\bm{W}^T\bm{X}_j\bm{W}))).
\label{Eq13}
\end{aligned}
\end{equation}

\setlength{\parskip}{1\baselineskip}

\noindent \textbf{Proposition 1.} \emph{The derivatives of the three trace forms $Tr(\cdot)$ in Eqn.\ref{Eq13} can be respectively computed by (Here, $\bm{B}_i= \bm{X}_i\bm{W}$, $\bm{\hat{X}}_i = \bm{W}^T\bm{X}_i\bm{W}$)}:
\begin{numcases}{}
D_W(Tr(\log^2(\bm{\hat{X}}_i)) = 4\bm{B}_i \text{D}\log(\bm{\hat{X}}_i)[\log(\bm{\hat{X}}_i)]. \label{Eq14} \\
D_W(Tr(\log^2(\bm{\hat{X}}_j)) = 4\bm{B}_j \text{D}\log(\bm{\hat{X}}_j)[\log(\bm{\hat{X}}_j)]. \label{Eq15} \\
\begin{aligned}
 & D_W(Tr(\log(\bm{\hat{X}}_i)\log(\bm{\hat{X}}_j)) \\ & = 2\bm{B}_i \text{D}\log(\bm{\hat{X}}_i)[\log(\bm{\hat{X}}_j)]
+2\bm{B}_j \text{D}\log(\bm{\hat{X}}_j)[\log(\bm{\hat{X}}_i)]. \label{Eq16}
\end{aligned}
\end{numcases}

\setlength{\parskip}{1\baselineskip}

\noindent \emph{Proof. The three formulas for the gradients with the matrix logarithm correspond to the three ones with rotation matrices in \cite{boumal2011discrete} (section 5.3), where a detailed proof is given.}

By using \textbf{Proposition 1.} (i.e. Eqn.\ref{Eq14}, Eqn.\ref{Eq15}, Eqn.\ref{Eq16}) and the sum rule of the directional derivatives, we derive $D_W k(\bm{W})$ with $\delta$ being the geodesic distance of LEM in Eqn.\ref{Eq4} as:
\begin{equation}
\begin{aligned}
D_W k_{ij}(\bm{W}) & = -4(\bm{B}_i \text{D}\log(\bm{\hat{X}}_i)[\log(\bm{\hat{X}}_i)-\log(\bm{\hat{X}}_j)] \\
 & +\bm{B}_j \text{D}\log(\bm{\hat{X}}_j)[\log(\bm{\hat{X}}_j)-\log(\bm{\hat{X}}_i)]) \beta k_{ij}(\bm{W}).
\label{Eq17}
\end{aligned}
\end{equation}

To calculate the formula Eqn.\ref{Eq17}, we then apply a function of block triangular matrix developed in \cite{al2009computing} to compute the form of $\text{D}\log(\bm{\hat{X}})[\bm{H}]$, which is the directional (Fr{\'e}chet) derivative of $\log$ at $\bm{\hat{X}} \in Sym^{+}_m$ along $\bm{H} \in Sym_{n}$. The following theorem shows that the directional derivative appears as the $(1,2)$ block of the resulting big matrix when $f: \bm{\hat{X}} \mapsto \log( \bm{\hat{X}})$ is evaluated at a certain block triangular matrix.

\setlength{\parskip}{1\baselineskip}

\noindent\textbf{Theorem 1.}  \emph{Let $f: \bm{\hat{X}} \mapsto \log( \bm{\hat{X}})$ be $2n-1$ times continuously differentiable on $\mathbb{G}$ and let the spectrum of $\bm{\hat{X}}$ lie in $\mathbb{G}$, where $\mathbb{G}$ is an open subset of $\mathbb{R}$. Then}
\begin{equation}
\begin{aligned}
f\left(\begin{bmatrix}
\bm{\hat{X}} & \bm{H}\\
0 & \bm{\hat{X}}\\
\end{bmatrix}\right)=\begin{bmatrix}
f(\bm{\hat{X}}) & \text{D}\log(\bm{\hat{X}})[\bm{H}]\\
0 & f(\bm{\hat{X}})\\
\end{bmatrix}.
\label{Eq18}
\end{aligned}
\end{equation}

\setlength{\parskip}{1\baselineskip}

\noindent\emph{Proof.} \emph{The result is proved by Najfeld and Havel \cite{najfeld1995derivatives} (Theorem 4.11) under the assumption that $f$ is analytic.}

By using \textbf{Theorem 1}, the directional derivative of the matrix logarithm can be easily computed. The pseudo matlab code of computing $\text{D}\log(\bm{\hat{X}})[\bm{H}]$ is simply listed as: n = size(X, 1);  Z = zeros(n); A = log([X, H ; Z, X]);  D = A(1:n, (n+1):end), where $\text{D}=\text{D}\log(\bm{\hat{X}})[\bm{H}]$.

With the derived gradient formulas in Eqn.\ref{Eq11}, Eqn.\ref{Eq19} and Eqn.\ref{Eq17}, the Euclidean gradient Eqn.\ref{Eq12} of the objective function Eqn.\ref{Eq9} for these cases can be computed to feed into the exploited RCG algorithm working on the PSD manifold. Since the global convergence of the RCG algorithm has been well-studied in the survey \cite{hager2006survey}, we do not investigate it any further here. The main time complexity of the algorithm is computing the gradient Eqn.\ref{Eq12}, being $O(lk_0n^2m+lk_1nm^2)$ ($l$ is the iteration number, $k_0/k_1$ is the number of selected samples/pairs, $n/m$ is the dimension of the original/target manifold) in the LEM case. In the experiment, we will also study the running time of each iteration of the algorithm by varying the number of selected between-class pairs for each SPD sample.

\section{Experiments}
\label{sec4}

In this section, we study the effectiveness of the proposed geometry-aware SPD similarity learning (SPDSL) approach by conducting experimental evaluations for three visual classification tasks including face recognition, material categorization and action recognition.

In these three tasks, the SPD features have been shown to provide powerful representations for images and videos via set-based covariances\cite{wang2012covariance,vemulapalli2013kernel,lu2013image}, region covariances \cite{tuzel2006region,tuzel2008pedestrian} and joint covariance descriptors \cite{Hussein2013,harandi2014manifold}. Therefore, they are natural choices to evaluate the proposed SPDSL exploiting Affine-Invariant metric (AIM), Stein divergence and Log-Euclidean metric (LEM).

To evaluate the effectiveness of the proposed SPDSL approach, we compare three categories of SPD-based learning methods, including basic Riemannian metric baseline methods, kernel learning based SPD discriminant learning methods and dimensionality reduction based SPD discriminant learning methods:

\begin{enumerate}
  \item Basic Riemannian metrics on SPD manifold:

   Affine-Invariant Metric (AIM) \cite{penec2006rieman}, Stein divergence \cite{sra2012new}, Log-Euclidean Metric (LEM) \cite{arsigny2007led}

  \item	Kernel learning based SPD matrix learning methods:

  PLS-based Covariance Discriminative Learning (CDL) \cite{wang2012covariance}, Riemannian Sparse Representation (RSR) \cite{harandi2012sparse} and Log-Euclidean Kernels (LEK) \cite{li2013log}

  \item Dimensionality reduction based SPD matrix learning methods:

    Log-Euclidean Metric Learning (LEML) \cite{huang2015leml} and SPD Manifold Learning (SPDML-AIM and SPDML-Stein) \cite{harandi2014manifold} with AIM and Stein divergence

\end{enumerate}
Note that, the proposed SPDSL belongs to the last category of SPD discriminant learning methods. As this paper focuses on studying the problem of supervised SPD discriminant learning, we here only report the performances of the original discriminant learning methods such as SPDML rather than those of further coupling them with other classifiers as done in the work \cite{harandi2014manifold}. In addition, in order to study the discriminant learning power of our proposed framework, we replace its supervised learning scheme with that of SPDML but still perform optimization on the exploited solution space. The adaption of the proposed SPDSL is denoted with SPDSL-AIM$^{*}$, SPDSL-Stein$^{*}$ and SPDSL-LEM$^{*}$.

%For most of the competing methods, we use the nearest neighbor classifier. The sparse representation based classifier is employed for the two sparse coding methods (i.e., RSR and LEK).

For RSR, the parameter $\beta$ was densely sampled around the order of the mean distance and the parameter $\lambda$ is sampled in the range of $[0.0001, 0.001, 0.01, 0.1]$. For LEK, there are three implements based on polynomial, exponential and radial basis kernels, which are respectively denoted as LEK-$\kappa_{p_n}$, LEK-$\kappa_{e_n}$ and LEK-$\kappa_g$. For LEK-$\kappa_{p_n}$ and LEK-$\kappa_{e_n}$, we densely sampled the $n$ from 1 to 50. The parameters $\beta$ in LEK-$\kappa_g$ and the $\lambda$ in the three LEK versions were all tuned in the same way as RSR. For LEML, the parameter $\eta$ is tuned in the range of [0.1, 1, 10], and $\zeta$ is tuned from 0.1 to 0.5. For SPDML and our method SPDSL, the maximum iteration number of the optimization algorithm is set to 50, the parameters $v_w$ is fixed as the minimum number of samples in one class, the dimensionality of the lower-dimensional SPD manifold and $v_b$ were tuned by cross-validation. The parameter $\beta$ in our method is set to $\beta=\frac{1}{\sigma^2}$, where $\sigma$ is equal to the mean distance of all pairs of training data.

%The dimensions of target manifolds for dimensionality reduction methods are all set as 40, 30 and 70 respectively on YTC, UIUC and HDM.

\begin{figure}[t]
\begin{center}
   \includegraphics[width=0.95\linewidth]{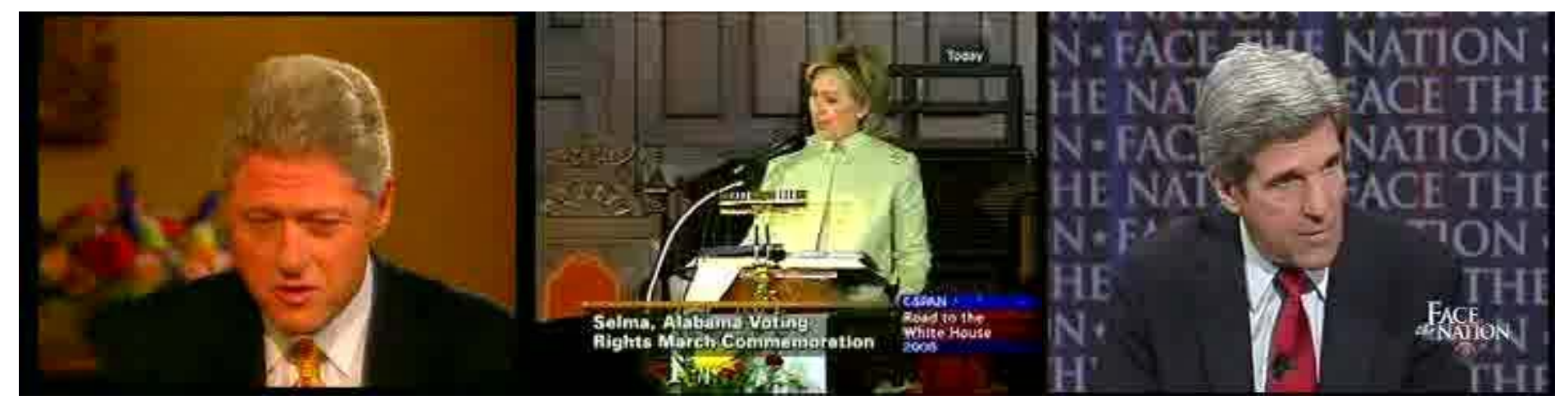}
\end{center}
   \caption{Video frames from the YTC video dataset \cite{kim2008ytc}.}
\label{fig:long}
\label{Fig3}
\end{figure}

\begin{table*}[t]
\linespread{1.2}
\caption{Average rank-1 face recognition rates (\%) with standard deviation of three categories of competing methods including the proposed SPDSL on the YTC database.}
\begin{center}
%\scriptsize
\begin{tabular}{|c|cccccc|}
\hline
%\noalign{\smallskip}
Category1 & AIM & Stein & LEM &  &  & \\
%\noalign{\smallskip}
\hline
Accuracy  & 62.85 $\pm$ 3.46 & 61.46 $\pm$ 3.52 & 63.91 $\pm$ 3.25 &  &  &   \\

\hline\hline
Category2 & CDL \cite{wang2012covariance} & RSR \cite{harandi2012sparse} & LEK-$\kappa_{p_n}$ \cite{li2013log} & LEK-$\kappa_{e_n}$ \cite{li2013log} &  LEK-$\kappa_g$ \cite{li2013log} &   \\
%\noalign{\smallskip}
\hline
Accuracy  & 72.67 $\pm$ 2.47 & 72.77 $\pm$ 2.69 & 61.85 $\pm$ 3.24 & 62.17 $\pm$ 3.52  & 56.30 $\pm$ 3.62 &   \\

\hline\hline
Category3 & LEML \cite{huang2015leml} & SPDML-AIM \cite{harandi2014manifold} & SPDML-Stein \cite{harandi2014manifold} & &  &  \\
\hline
Accuracy  & 70.53 $\pm$ 2.95 & 64.66 $\pm$ 2.92 & 61.57 $\pm$ 3.43 &  & &  \\

\hline
The proposed &  \textbf{SPDML-AIM$^{*}$}  & \textbf{SPDML-Stein$^{*}$} & \textbf{SPDML-LEM$^{*}$} & \textbf{SPDSL-AIM} & \textbf{SPDSL-Stein} &  \textbf{SPDSL-LEM} \\
\hline
Accuracy  & \textbf{64.27 $\pm$ 2.84}  & \textbf{62.31  $\pm$ 3.48} & \textbf{69.32 $\pm$ 2.04} & \textbf{71.60 $\pm$ 2.45} & \textbf{71.03  $\pm$ 2.39} & \textbf{72.29 $\pm$ 1.58}\\
\hline
\end{tabular}
\end{center}
\label{tab2}
\end{table*}

\subsection{Face Recognition}

As the first experiment, we use YouTube Celebrities (YTC) video face database \cite{kim2008ytc} to perform the task of video face recognition. The dataset is a quite challenging and widely used in the study of video face recognition. It has 1,910 video clips of 47 subjects collected from YouTube. Most clips contains hundreds of frames, which are often low resolution and highly compressed with noise and low quality.

For the testing protocol, following \cite{wang2012covariance,vemulapalli2013kernel,huang2015leml}, this dataset is randomly split into the gallery and the probe, which have 3 image sets and 6 image sets respectively for each subject. The process of random testing was repeated 10 times for the evaluation on video face recognition.

In our experiment, each face image in videos is cropped into $20 \times 20$ intensity image and then histogram equalized to eliminate lighting effects. Following the works \cite{wang2012covariance, huang2015leml}, we extract the set-based covariance matrix for each video sequence of frames on this dataset. To avoid matrix singularity, we add a small ridge $\delta \bm{I}$ to each covariance matrix $\bm{\Sigma}$, where $\delta=10^{-3} \times trace(\bm{\Sigma})$ and $\bm{I}$ is the identity matrix. In the literature, the mean face in each video has been proved to benefit video face recognition. Therefore, we improve the set-based covariance matrix feature by concatenating it with the mean to yield a $(d+1)$-dimensional SPD matrix as $\begin{bmatrix}
    \bm{\Sigma}+\bm{{\mu}{\mu}}^T & \bm{{\mu}}\\
    \bm{\mu}^T & 1\\
    \end{bmatrix}$, where  $\bm{{\mu}} \in \mathbb{R}^d$ and $\bm{\Sigma} \in \mathbb{S}_{+}^d$ represents the mean and the covariance matrix of one image set. Note that the dimensions of target manifolds for dimensionality reduction methods are all set as 40 for the YTC database.

As can be seen from Table \ref{tab2}, the baseline method LEM outperforms the other two baselines AIM and Stein in most of cases, which demonstrates that the LEM is more effective than the other two Riemannian metrics in the evaluation. The results in Table \ref{tab2} also show that most of the kernel learning based (Category2) and dimensionality reduction based (Category3) methods boost the accuracies of the baselines AIM, Stein and LEM. This also concludes that learning discriminative SPD features in these methods can help the visual recognition tasks.

Compared with the state-of-the-art kernel learning based methods CDL and RSR, the dimensionality reduction based methods LEML and SPDML perform worse in the task. In contrast, our SPDSL improves LEML and SPDML by around 2\% and 7\% respectively, and achieve comparable performance with CDL and RSR.
In the comparison with SPDML, the performances of the adaption of our new SPD similarity learning framework SPDML-AIM$^{*}$ and SPDML-Stein$^{*}$ are close to those of SPDML-AIM and SPDML-Stein. This shows the former solution can be approximated by the latter solution when the involved Riemannian metric is affine invariant. \mbox{Nevertheless}, after using the proposed supervised learning technique, SPDSL-AIM and SPDSL-Stein clearly outperform SPDML method. In addition, our SPDSL can handle the case when the SPD manifold is equipped with the non-affine Riemannian metric LEM, and we see that SPDML-LEM$^{*}$ and SPDSL-LEM achieve higher accuracies in most cases.

%, which shows SPDSL is an effective supervised SPD similarity learning method
% which shows the superiority of our method over SPDML in the term of the ability of discriminant learning on SPD feature

\begin{figure}[t]
\begin{center}
   \includegraphics[width=0.95\linewidth]{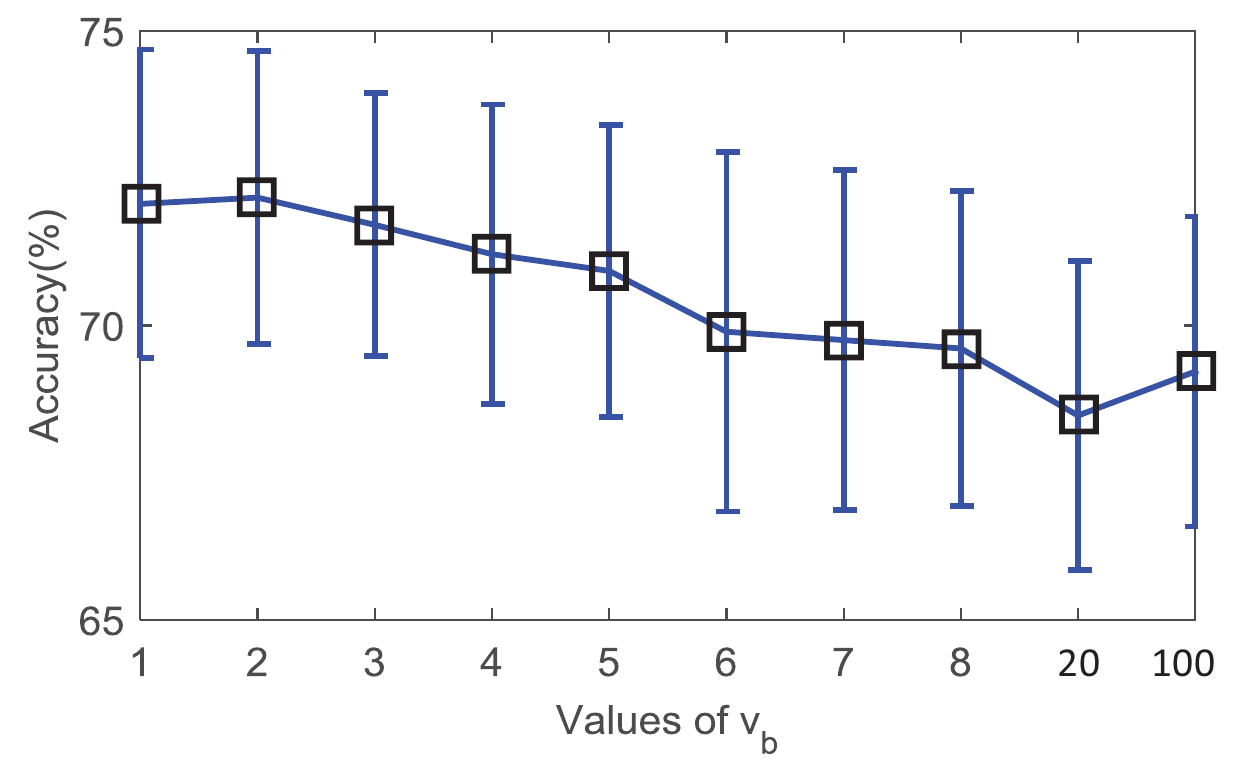}
\end{center}
   \caption{Recognition accuracy of the proposed SPDSL-LEM on the YTC dataset for varying values of $v_b$ (i.e., different sparse degrees of the involved kernel matrix $k({\bm{W}})$).}
\label{fig:long}
\label{Fig6}
\end{figure}
\vspace{1em}

Besides, we study the effectiveness of the proposed SPDSL technique when varying its key parameter $v_b$. As shown in Fig.\ref{Fig6}, we present the behavior of the sparse (non PD) kernel cases on the YTC database for different values of $v_b$ in the interval $[1,8]$ and the values of 20 and 100 while fixing the parameter $v_w=3$. When $k({\bm{W}})$ achieves a full kernel matrix, the performance gets to 72.57\%, which is close to the highest performance (72.29\%) reached by the sparse kernel matrix cases (see Fig.\ref{Fig6}).

%The intra-class discrimination plays the dominant role when $v_b=1$, while $v_b=100$ makes the inter-class discrimination dominant. The highest performance is reached when $v_b= 2$, which demonstrates that balance between the intra-class and inter-class terms is important.
%The full (PD for LEM) kernel case (i.e., $v_b$ reaches the largest and ),

%\textcolor[rgb]{0.00,0.00,1.00}{Additionally, on UIUC, HDM and YTC, the mean accuracies (52.13\%, 49.13\%, 72.29\%, reported in Tab.2) of the sparse kernel case are very close to those (51.94\%, 48.88\%, 72.57\%) of the full (PD in the case of LEM) kernel case.}

\begin{figure}[t]
\begin{center}
   \includegraphics[width=0.95\linewidth]{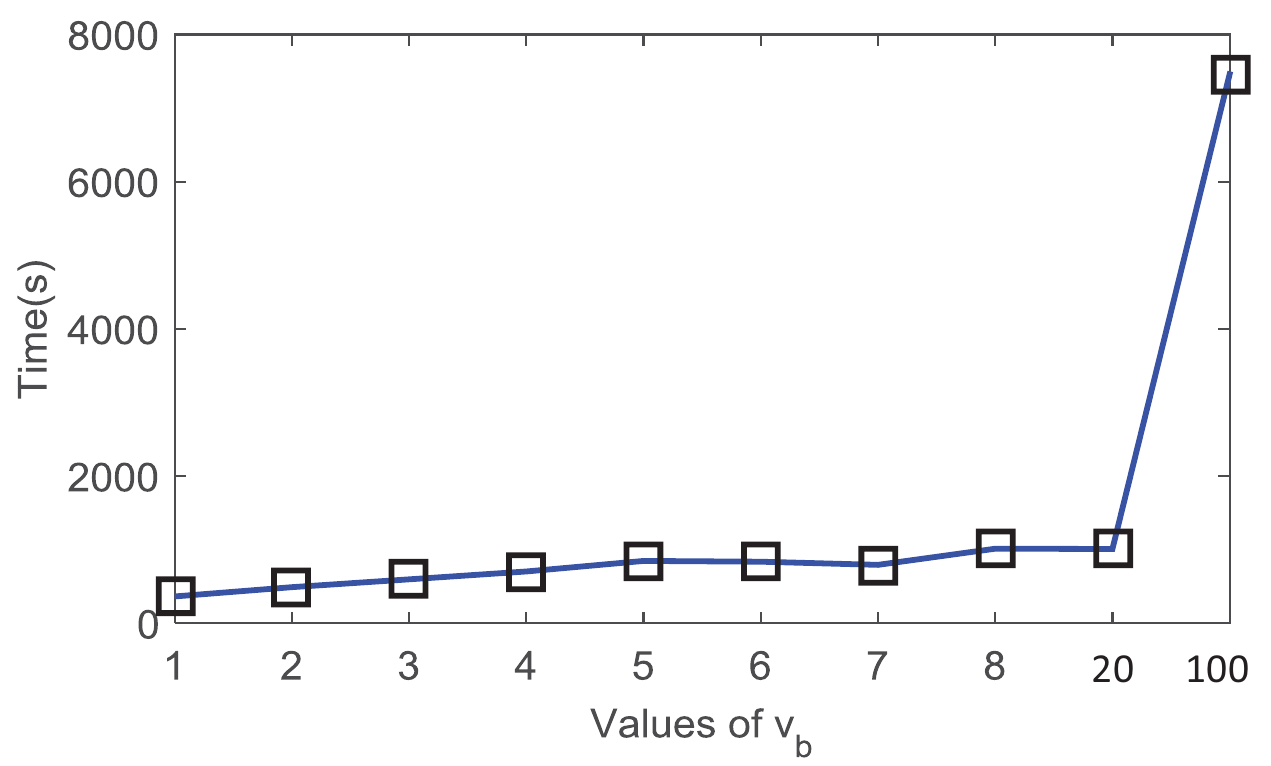}
\end{center}
   \caption{Running time of the proposed SPDSL-LEM on the YTC dataset for varying values of $v_b$ (i.e., different sparse degrees of the involved kernel matrix $k({\bm{W}})$).}
\label{fig:long}
\label{Fig7}
\end{figure}
\vspace{1em}

The efficiency of the proposed SPDSL technique is studied as well. As shown in Fig.\ref{Fig7}, the running time is average training time of each iteration of the optimization algorithm, which typically iterates 50 times. Specifically, we perform the testing on the YTC dataset, and employ an Intel(R) Core(TM) i5-2400 (3.10GHz) PC. As the value of $v_b$ increases, the running time turns to be much higher especially when $k({\bm{W}})$ is full, whose running time is around 13,975 seconds (i.e., about 30 times of the case of $v_b= 2$ at each iteration, and extremely expensive when the algorithm iterates 50 times) running on YTC. Hence, when huge datasets are involved, the sparse kernel case scales much better than the full (PD) kernel case with very slight gain/loss of accuracy.

\begin{figure}[t]
\begin{center}
   \includegraphics[width=0.95\linewidth]{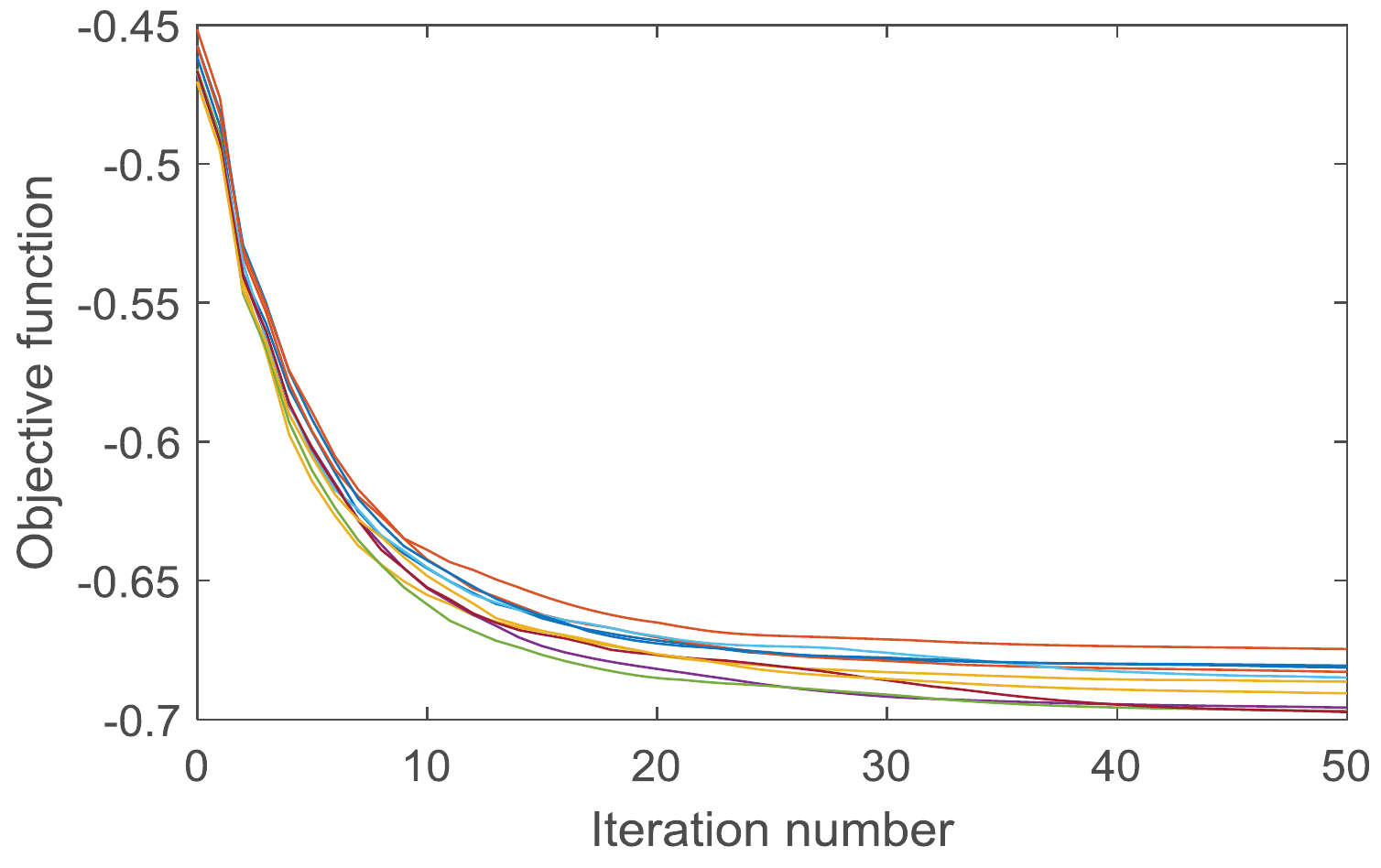}
\end{center}
   \caption{Convergence behavior of the exploited RCG algorithm for SPDSL-LEM in 10 random testings of the YTC dataset with the parameter $v_b=2$.}
\label{fig:long}
\label{Fig8}
\end{figure}
\vspace{1em}

In the end, we also investigate the convergence behavior of the exploited RCG algorithm for our SPDSL approach. As seen from the results in Fig.\ref{Fig8}, the optimization algorithm on the PSD manifold is able to converge to a favorable solution after several tens of iterations.

\begin{figure}[t]
\begin{center}
   \includegraphics[width=0.95\linewidth]{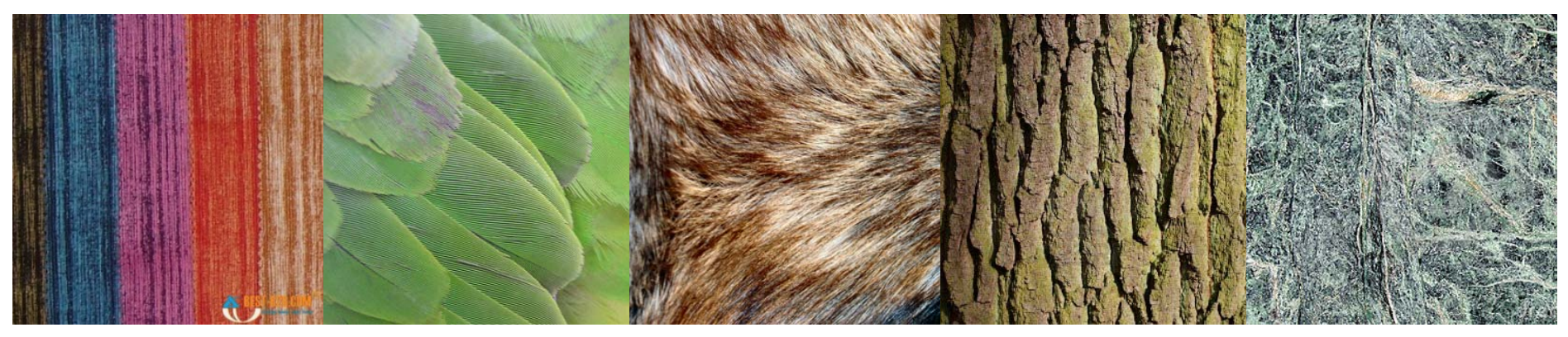}
\end{center}
   \caption{Samples from the UIUC material dataset \cite{liao2013nonpara}.}
\label{fig:long}
\label{Fig4}
\end{figure}

\begin{table*}[t]
\linespread{1.2}
\caption{Average recognition accuracies (\%) with standard deviation of three categories of competing methods including the proposed SPDSL on the UIUC database.}
\begin{center}
%\scriptsize
\begin{tabular}{|c|cccccc|}
\hline
%\noalign{\smallskip}
Category1 & AIM & Stein & LEM &  &  & \\
%\noalign{\smallskip}
\hline
Accuracy  & 46.30 $\pm$ 2.90 & 42.87 $\pm$ 2.27 & 46.30 $\pm$ 2.86 &  &  &   \\

\hline\hline
Category2 & CDL \cite{wang2012covariance} & RSR \cite{harandi2012sparse} & LEK-$\kappa_{p_n}$ \cite{li2013log} & LEK-$\kappa_{e_n}$ \cite{li2013log} &  LEK-$\kappa_g$ \cite{li2013log} &   \\
%\noalign{\smallskip}
\hline
Accuracy  & 54.91 $\pm$ 4.72 & 52.41 $\pm$ 4.03 & 48.89 $\pm$ 3.29 & 49.54 $\pm$ 3.67  & 49.63 $\pm$ 3.03 &   \\

\hline\hline
Category3 & LEML \cite{huang2015leml} & SPDML-AIM \cite{harandi2014manifold} & SPDML-Stein \cite{harandi2014manifold} & &  &  \\
\hline
Accuracy  & 52.53 $\pm$ 2.13 & 48.09 $\pm$ 1.82 & 49.17 $\pm$ 2.37 &  & &  \\

\hline
The proposed &  \textbf{SPDML-AIM$^{*}$}  & \textbf{SPDML-Stein$^{*}$} & \textbf{SPDML-LEM$^{*}$} & \textbf{SPDSL-AIM} & \textbf{SPDSL-Stein} &  \textbf{SPDSL-LEM} \\
\hline
Accuracy  & \textbf{50.00 $\pm$ 3.60} & \textbf{49.35 $\pm$ 2.47} & \textbf{50.28 $\pm$ 3.78} & \textbf{52.31 $\pm$ 3.55} & \textbf{51.57 $\pm$ 4.16} & \textbf{52.13 $\pm$ 3.49} \\
\hline
\end{tabular}
\end{center}
\label{tab3}
\end{table*}

\subsection{Material Categorization}

For the task of material categorization, we conduct experiments on the UIUC material dataset \cite{liao2013nonpara}. This dataset includes 18 subcategories of materials taken in the wild from four general categories: \emph{bark, fabric, construction materials}, and \emph{outer coat of animals}. Each subcategory contains 12 images taken at different scales. Several samples from this database are shown in Fig.\ref{Fig4}.

Both Region Covariance Matrices (RCMs) \cite{tuzel2006region} and SIFT features \cite{lowe2005ijcv} have been shown to be robust and discriminative for material categorization \cite{liao2013nonpara}. As done in \cite{harandi2014manifold}, we extract RCMs of size $128\times 128$ using 128 dimensional SIFT features from gray scale images. Specifically, we resize each image to $400 \times 400$ and compute the dense SIFT descriptors on a grid with 4 pixels spacing (each patch size is 16x16, the number of angles is 8, the number of Bins is 4). In each grid point, one 128-dimensional SIFT feature is thus yielded. For dimensionality reduction methods, the dimensions of target manifolds are all set as 40 in the evaluation.

Following the work \cite{harandi2014manifold}, on the UIUC dataset, we randomly select half of the images from each subcategory as training data, and the remaining images as testing data. This process of evaluation is conducted 10 times in our experiment.

In Table \ref{tab2}, for the competing methods, we report their average accuracies with standard deviations of 10 random testings on the UIUC dataset. As concluded in the last evaluation, the proposed dimensionality reduction technique SPDSL improves the most related method SPDML method by 2\%-4\%, and achieves comparable performances with the state-of-the-art methods.

%Note that, on the database, we only extracted dense SIFT to construct SPD features while the work \cite{harandi2014manifold} fused the dense SIFT and color feature.

\begin{figure}[t]
\begin{center}
   \includegraphics[width=0.95\linewidth]{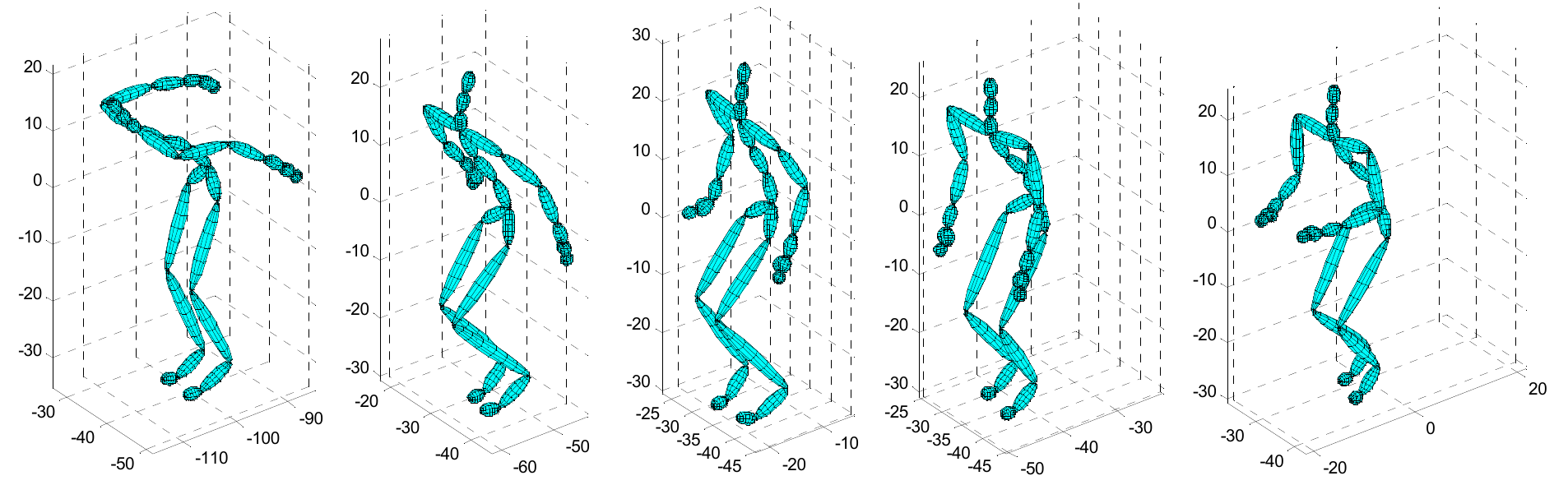}
\end{center}
   \caption{Hopping action from the HDM05 Motion Capture database \cite{muller2007hdm}.}
\label{fig:long}
\label{Fig3}
\end{figure}

\begin{table*}[t]
\linespread{1.2}
\caption{Average recognition accuracies (\%) with standard deviation of three categories of competing methods including the proposed SPDSL on the HDM05 database.}
\begin{center}
%\scriptsize
\begin{tabular}{|c|cccccc|}
\hline
%\noalign{\smallskip}
Category1 & AIM & Stein & LEM &  &  & \\
%\noalign{\smallskip}
\hline
Accuracy  & 42.70 $\pm$ 1.74 & 42.13 $\pm$ 2.63 & 43.98 $\pm$ 2.13 &  &  &   \\

\hline\hline
Category2 & CDL \cite{wang2012covariance} & RSR \cite{harandi2012sparse} & LEK-$\kappa_{p_n}$ \cite{li2013log} & LEK-$\kappa_{e_n}$ \cite{li2013log} &  LEK-$\kappa_g$ \cite{li2013log} &   \\
%\noalign{\smallskip}
\hline
Accuracy  & 41.74 $\pm$ 1.92 & 41.12 $\pm$ 2.53  & 47.22 $\pm$ 1.62 & 46.87 $\pm$ 1.72  & 48.72 $\pm$ 3.00 &   \\

\hline\hline
Category3 & LEML \cite{huang2015leml} & SPDML-AIM \cite{harandi2014manifold} & SPDML-Stein \cite{harandi2014manifold} & &  &  \\
\hline
Accuracy  & 46.87 $\pm$ 2.19 & 47.25 $\pm$ 2.78 & 46.21 $\pm$ 2.65 &  & &  \\

\hline
The proposed &  \textbf{SPDML-AIM$^{*}$}  & \textbf{SPDML-Stein$^{*}$} & \textbf{SPDML-LEM$^{*}$} & \textbf{SPDSL-AIM} & \textbf{SPDSL-Stein} &  \textbf{SPDSL-LEM} \\
\hline
Accuracy  & \textbf{47.93 $\pm$ 2.62} & \textbf{46.35 $\pm$ 2.45} & \textbf{48.88 $\pm$ 3.18} & \textbf{48.09 $\pm$ 2.49} & \textbf{49.02 $\pm$ 2.93}  & \textbf{49.13 $\pm$ 2.74} \\
\hline
\end{tabular}
\end{center}
\label{tab4}
\end{table*}

\subsection{Action Recognition}

We then employ the HDM05 database \cite{muller2007hdm} to handle with the problem of human action recognition from motion capture sequences. As shown in Fig.\ref{Fig3}, this dataset contains 2,337 sequences of 130 motion classes, e.g., \emph{`clap above head',`lie down floor',`rotate arms', `throw basket ball'}, in 10 to 50 realizations executed by various actors.

The 3D locations of 31 joints of the subjects are provided over time acquired at the speed of 120 frames per second. Following the previous works \cite{Hussein2013,harandi2014manifold}, we represent an action of a $K$ joints skeleton observed over m frames by its joint covariance descriptor. This descriptor is an form of SPD matrix of size $3K \times 3K$, which is computed by the second order statistics of 93-dimensional vectors concatenating the 3D coordinates of the 31 joints in each frame.

As the evaluation protocol on UIUC, on this dataset, we also conduct 10 times random evaluations, in which half of sequences (around 1,100 sequences) are randomly selected for training data, and the rest are used for testing. On the HDM05 database, the work \cite{harandi2014manifold} only used 14 motion classes for evaluation while we tested these methods for identifying 130 action classes. %Accordingly, our reported recognition rates of the related methods are a bit lower than those published in \cite{harandi2014manifold}.

Table.\ref{tab4} summarizes the performances of the comparative algorithms on the UIUC dataset. In the evaluation, the dimensions of resulting manifolds achieved by dimensionality reduction methods are all set as 30. Different from the last two evaluations, CDL and RSR performance worse than other competing methods. The proposed SPDSL again improves the existing dimensionality reduction based methods LEML and SPDML with 1\%-3\%, and achieve state-of-the-art performance on the HDM05 database.

%We can see that our SPDNet outperforms the state-of-the-art shallow SPD matrix learning methods by a large margin (more than 13\%). This shows that the proposed non-linear deep learning scheme on SPD matrices leads to great improvements when the training data is large enough.

\subsection{Discussion}

Since our method SPDSL and the two methods SPDML, LEML adopt the same SPD matrix learning scheme, we here mainly make two pieces of discussions between them.

First, compared with the related manifold learning method SPDML, our SPDSL framework proposes a more general solution and a more favorable objective function. This point has been validated by the three evaluations. As can be seen from Table \ref{tab2}, Table \ref{tab3} and Table \ref{tab4}, there are two key conclusions observed from the three visual recognition tasks:

a) As for the new solution, its main benefits lie in enlarging the search domain and opening up the possibility of using non-affine invariant metrics (e.g. LEM). While SPDML* for affine invariant metrics AIM and Stein improves SPDML mildly (this may depend on the data), the gains of SPDML*-LEM over the AIM and Stein cases are relatively obvious, i.e. 1.65\%, 2.15\%, 6.21\% on average, respectively for the three datasets.

b) The new objective function (for similarity regression) is quite different from that (for graph embedding) used in [5]. While it's hard to theoretically prove the gains, we have empirically studied its priority. By comparing SPDSL with SPDML*, the improvements for the three datasets are 2.13\%, 1.03\%, 6.34\% on average for the three used databases, respectively.

Second, in contrast to LEML which focuses on metric learning, our SPDSL learns discriminative similarities on SPD manifolds. Besides, while LEML performs metric learning on the tangent space of SPD manifolds, the proposed SPDSL learns similarity directly on the SPD manifolds. Intuitively, our learning scheme would more faithfully respect the Riemannian geometry of the data space, and thus could lead to more favorable SPD features for classification tasks. From the above three evaluations, we can see some improvements of SPDSL over LEML.

%------------------------------------------------------------------------
\section{Conclusions} \label{sec5}

We have proposed a geometry-aware SPD similarity learning (SPDSL) framework for more robust visual classification tasks. Under this framework, by exploiting the Riemannian geometry of PSD manifolds, we open the possibility of directly learning the manifold-manifold transformation matrix. To achieve the discriminant learning on the SPD features, this work devises a new SPDSL technique working on SPD manifolds. With the objective of the proposed SPDSL, we derive an optimization algorithm on PSD manifolds to pursue the transformation matrix. Extensive evaluations have studied both the effectiveness of efficiency of our SPDSL on three challenging datasets.

%, which expands the problem domain of SPD similarity learning

For future work, the study on the relationship between the selected Riemannian metrics of PSD manifolds and SPD manifolds would be interesting for the problem of supervised SPD similarity learning. Besides, if neglecting the designed discriminant function on SPD features, learning the transformation on SPD features for object sets is equal to learning the projection on single object features. Thus, this work can be extended to learn hierarchical representations on object feature by leveraging the current powerful deep learning techniques.

%\appendices
%\section{Proof of the First Zonklar Equation}
%Appendix one text goes here.
%
%% you can choose not to have a title for an appendix
%% if you want by leaving the argument blank
%\section{}
%Appendix two text goes here.

% use section* for acknowledgment
\section*{Acknowledgment}

This work has been carried out mainly at the Institute of Computing Technology (ICT), Chinese Academy of Sciences (CAS). It is partially supported by 973 Program under contract No. 2015CB351802, Natural Science Foundation of China under contracts Nos. 61390511, 61173065, 61222211, and 61379083.

%The authors would like to thank...

% Can use something like this to put references on a page
% by themselves when using endfloat and the captionsoff option.
\ifCLASSOPTIONcaptionsoff
  \newpage
\fi

\end{document}